\documentclass[11pt, a4paper,logo,  copyright, nonumbering]{paddleocr}
\usepackage[numbers]{natbib}
\usepackage{dblfloatfix}
\usepackage{ulem}
\usepackage{float}
\usepackage{tocloft}
\usepackage{caption}
\usepackage{dramatist}
\usepackage{xspace}
\usepackage{array}
\usepackage{pifont}
\usepackage{multirow}
\usepackage{geometry}
\usepackage{makecell}
\usepackage{tcolorbox}
\usepackage{xltabular}
\usepackage{titlesec}
\usepackage{longtable}
\usepackage[hang, flushmargin]{footmisc}
\usepackage{booktabs}
\usepackage{xcolor}
\usepackage[table]{xcolor}
\definecolor{lightpink}{RGB}{255, 182, 193}
\definecolor{iccvblue}{rgb}{0.21,0.49,0.74}

\hypersetup{
    colorlinks=true,
    linkcolor=blue,
    citecolor=blue,
    filecolor=blue,
    urlcolor=blue,
    pdfpagemode=UseNone,
    pdfstartview=FitH
}
\usepackage{threeparttable}
\usepackage{soul}
\usepackage{amsfonts}
\usepackage{amsmath}
\usepackage{amssymb}
\usepackage{lineno}
\usepackage{adjustbox}
\usepackage{CJKutf8}
\usepackage{subcaption}
\usepackage{setspace}
\usepackage{dsfont}
\usepackage{tabularx}
\usepackage{lipsum}
\usepackage{multicol}
\usepackage{listings}
\usepackage{tablefootnote}
\usepackage{hyperref}
\usepackage{bm}
\usepackage{tikz}
\usepackage{comment}
\usepackage{enumitem}

\makeatletter
\def\@BTrule[#1]{%
  \ifx\longtable\undefined
    \let\@BTswitch\@BTnormal
  \else\ifx\hline\LT@hline
    \nobreak
    \let\@BTswitch\@BLTrule
  \else
     \let\@BTswitch\@BTnormal
  \fi\fi
  \global\@thisrulewidth=#1\relax
  \ifnum\@thisruleclass=\tw@\vskip\@aboverulesep\else
  \ifnum\@lastruleclass=\z@\vskip\@aboverulesep\else
  \ifnum\@lastruleclass=\@ne\vskip\doublerulesep\fi\fi\fi
  \@BTswitch}
\makeatother

\addto\extrasenglish{
}

 {\begin{list}{}%
         {\setlength{\leftmargin}{#1}}%
         \item[]%
 }
 {\end{list}}

\reportnumber{001}

\renewcommand{\today}{}

\title{\centering P-MTP: Efficient Document Parsing via Multi-Token Prediction with Progressive Depth Scaling}

\author[*]{
\small
Le Xiang\textsuperscript{1,*}, Chenxi Zhai\textsuperscript{2,*}, Shu Wei\textsuperscript{1,\dag},
Jingjing Wu\textsuperscript{1}, Qunyi Xie\textsuperscript{1,\ddag}, Xiao Tan\textsuperscript{1}, Kunbin Chen\textsuperscript{1}, Wei He\textsuperscript{1}
\vspace{-0.4cm}\\
\small
\textsuperscript{*}Equal contribution. \textsuperscript{\dag}Corresponding author. \textsuperscript{\ddag}Team leader
\vspace{0.2cm}\\
\small
\textsuperscript{1}Department of Computer Vision Technology (VIS), Baidu Inc., China\\
\small
\textsuperscript{2}Tsinghua University, Shenzhen International Graduate School, China\\
\small
\texttt{\{xiangle, weishu01, wujingjing06, xiequnyi, tanxiao01, chenkunbin, hewei06\}@baidu.com}\\
\small
\texttt{\{dcx24\}@mails.tsinghua.edu.cn}
\vspace{0.2cm}\\
{\small
\includegraphics[height=0.9em]{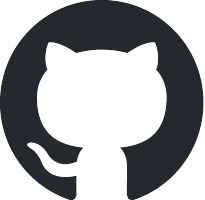} \textbf{Source Code}: \url{https://github.com/Hor1zonz/PMTP}
}
}

\renewcommand{\phi}{\varphi}

\renewcommand{\leq}{\leqslant}

\renewcommand{\epsilon}{\varepsilon}
\renewcommand{\imath}{\mathrm{i}}

\newlength{\restsubwidth}
\newlength{\restsubheight}
\newlength{\restsubmoreheight}
\newcommand{\rest}[2]{%
        \settowidth{\restsubwidth}{\ensuremath{#2}}
        \settoheight{\restsubheight}{\ensuremath{{}_{#2}}}
        \ensuremath{{#1\hskip 0.5pt}_{\vrule\kern2pt\parbox[b][%
        4pt][b]{\the\restsubwidth}{%
                        \ensuremath{{}_{#2}}}}}
        }

\definecolor{myGreen}{RGB}{169, 209, 142}
\definecolor{shadecolor}{rgb}{0.9, 0.9, 0.9}

\newcommand{\wu}[1]{\textcolor[rgb]{0,0,0}{#1}}
\newcommand{\xie}[1]{\textcolor[rgb]{0,0,0}{#1}}
\newcommand{\mymodel}{P-MTP}

\fancypagestyle{firststyle}{
    \fancyhead[L]{
        \ifthenelse{\boolean{logo}}{
            \includegraphics[width=120pt]{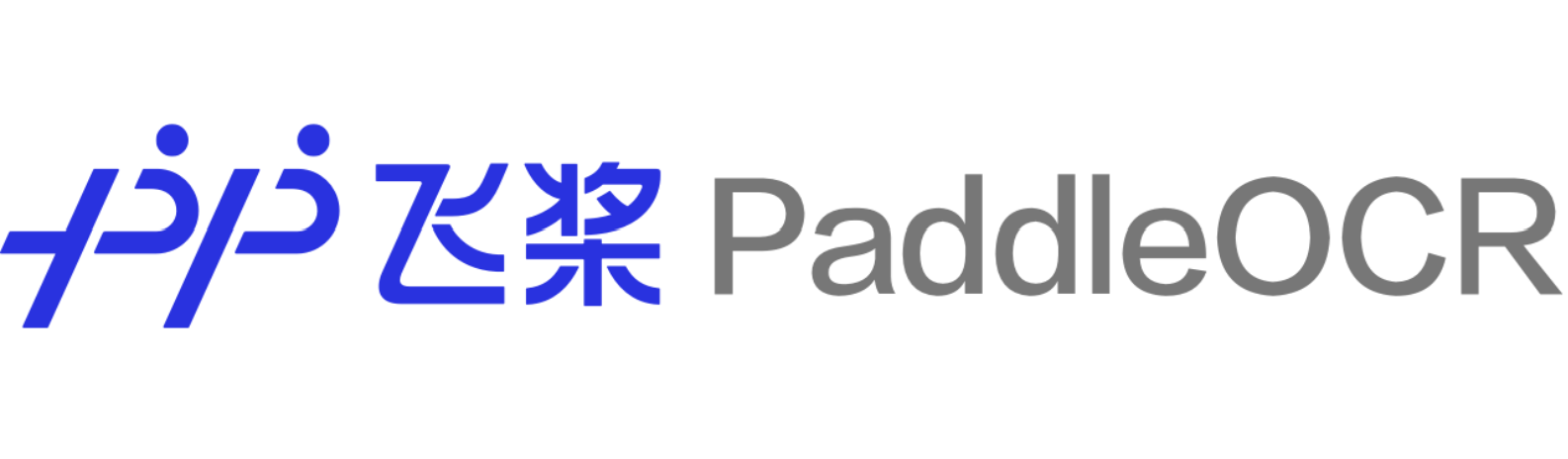}
        }{}
        \ifdefined\paperurl
        \if\relax\the\paperurl\relax \else
            \href{\the\paperurl}{\urlheaderfont \itshape \the\paperurl}\\ \fi
        \else \fi
        {\footerfont\itshape\monthyeardate\today}
    }
    \fancyhead[R]{
       \raisebox{-5pt}{\includegraphics[height=45pt]{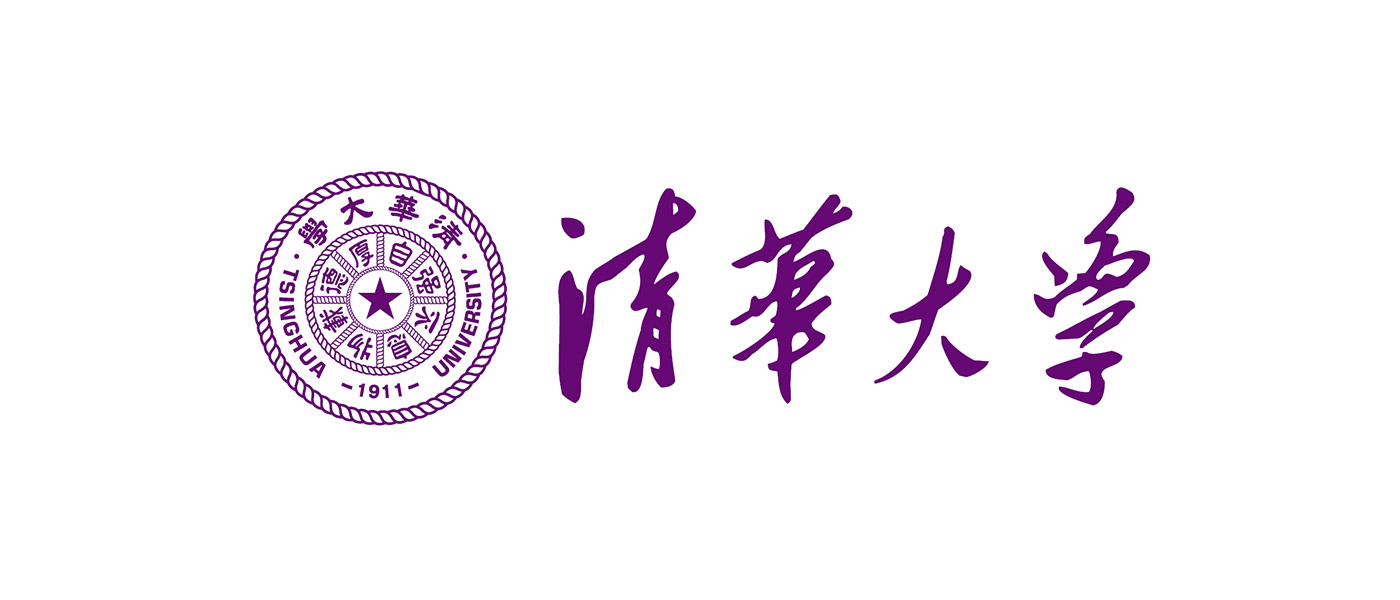}}   
    }
    \fancyhead[C]{}
    \fancyfoot[L]{
    	\ifdefined\correspondingauthor
    	\if\relax\the\correspondingauthor\relax
    	\else \footerfont {*\the\correspondingauthor \\} \fi
    	\else \fi
        \ifthenelse{\boolean{internal}}{\footerfont \internalonly \\}{\footerfont\bfseries\relax}
    	\ifthenelse{\boolean{address}}{
    		\itshape\footerfont PaddlePaddle Team, Baidu Inc. \\}{}
    }
    \fancyfoot[R]{
    	\ifthenelse{\boolean{internal}}{
    	\ifdefined\reportnumber
    	\if\relax\the\reportnumber\relax
    	\else \footerfont\itshape  {\footerfont \bfseries

    	} \fi
    	\else \fi
    	}{\footerfont\bfseries\relax}
    }
    \fancyfoot[C]{\footerfont\bfseries\relax}
}

\begin{abstract}
Vision-Language Models (VLMs) have revolutionized document parsing by enabling end-to-end mapping from images to structured text, imposing a significant latency bottleneck, particularly for token-dense documents. While Multi-Token Prediction (MTP) has emerged as a promising approach for accelerating inference, its potential is constrained by optimization instability when scaling to deeper look-ahead depth. In this paper, we propose \textbf{\mymodel}, a framework that leverages \textbf{Progressive Multi-Token Prediction} with a lightweight MTP module to scale the look-ahead depth for high-throughput document parsing. Specifically, we introduce Progressive Curriculum Loss that adaptively re-weights different look-ahead depths using cumulative path reliability and retrospective target consistency. By effectively suppressing gradient noise in long-range predictions, \mymodel\, facilitates an automated easy-to-hard optimization transition, enabling the model to master increasingly distant look-ahead depths. Furthermore, we propose Confidence-Gated Dynamic Drafting to maximize the effective look-ahead depth and acceptance rate by adaptively calibrating speculative length during inference, thereby minimizing computational waste and further pushing the boundaries of inference speedup. Experimental results across multiple benchmarks and architectures demonstrate that \mymodel\, achieves up to a $5\times$ speedup with negligible loss in accuracy, providing the first successful validation of extensive look-ahead MTP in the document parsing domain.
\end{abstract}

\begin{document}
\maketitle 
\section{Introduction}
\label{sec:intro}

Document parsing refers to the automated transformation of unstructured document images into structured representations such as JSON, Markdown, or LaTeX~\cite{zhang2024document}. As a fundamental component in the modern AI-driven data pipeline, this task is crucial for enabling high-throughput information extraction and facilitating the seamless integration of physical documents into large-scale knowledge bases.

The autoregressive decoding paradigm has emerged as the dominant framework for cross-modal generation. While Vision-Language Models (VLMs) have established state-of-the-art performance in document parsing, their step-by-step nature incurs prohibitive latency for token-dense layouts. To mitigate this, existing works primarily focus on structural task fragmentation~\cite{cui2025paddleocr,wang2024mineru,yin2026youtu} or architectural compression~\cite{feng2025dolphin,feng2026dolphin2}. Recent explorations have shifted toward decoding-level acceleration~\cite{yin2026youtu,GLM-OCR} to alleviate the latency of token-dense document parsing.

\begin{figure}[t]
    \centering
    \includegraphics[width=1\linewidth]{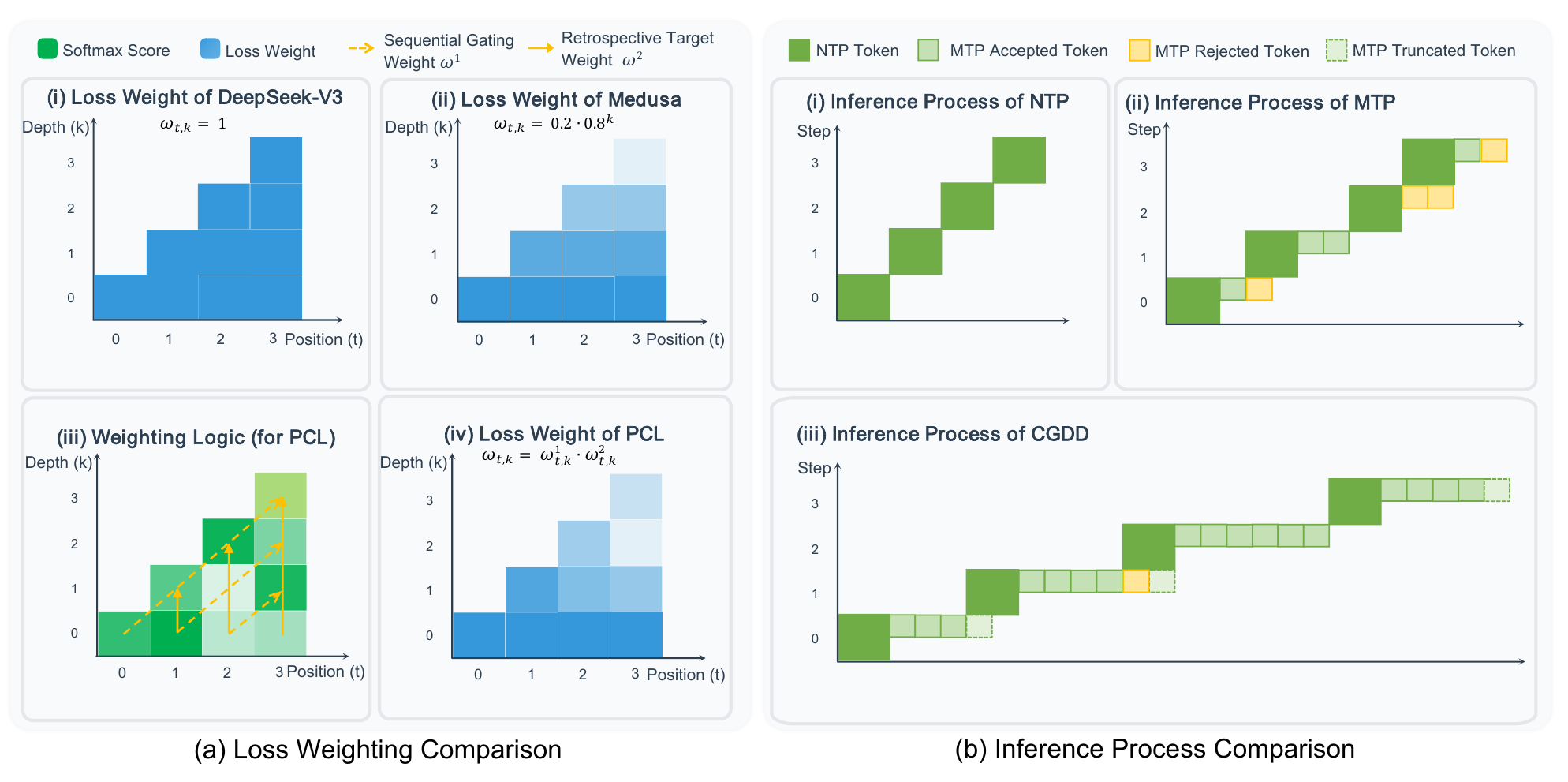}
    \caption{Comparative illustration of P-MTP training and inference, where (a) shows a comparison of different MTP weighting schemes and (b) shows a comparison of drafting paradigms. In (a), unlike the constant weights in (i) DeepSeek-V3 or static decay in (ii) Medusa, P-MTP introduce Progressive Curriculum Loss which weights look-ahead depths adaptively via ($\omega^1, \omega^2$). In (b), compared to (i) NTP and (ii) fixed-depth MTP, (iii) P-MTP adaptively extends the drafting depth using confidence-gating, maximizing accepted tokens while truncating unreliable paths.}
    \label{fig:teaser}
\end{figure}

As a perception-heavy task, document parsing produces text sequences that are tightly coupled with invariant visual features. This results in a high-certainty predictive distribution, making it an ideal candidate for Multi-Token Prediction (MTP) where such confidence theoretically enables a deeper look-ahead reach. Prior MTP works in the LLM field ~\cite{ankner2024hydra,li2025amphista,li2025gumiho,zhang2024draft,zeng2025resdecode} has achieved significant performance acceleration. However, scaling the look-ahead depth in document domain remains constrained by the inherent limitations of existing training and inference paradigms. As illustrated in Figure ~\ref{fig:teaser} (a), existing methods typically employ constant or static-decay weighting schemes that remain agnostic to trajectory divergence. This often leads to training instability and hallucinated gradients when distal supervision is applied despite proximal prediction failures. Furthermore, as shown in Figure ~\ref{fig:teaser} (b), traditional MTP relies on fixed-depth drafting~\cite{chen2023accelerating}, which results in frequent rejections in complex scenarios and underutilizes the look-ahead potential in simple patterns.

To overcome these limitations and address the aforementioned research gap, we introduce P-MTP (Progressive Multi-Token Prediction), which integrates a Progressive Curriculum Loss to stably scale the look-ahead depth. This mechanism utilizes an adaptive loss weight $\omega_{t,k}$ comprised of Sequential Confidence Constraint ($\omega^1$) and Retrospective Target Constraint ($\omega^2$) to modulate the supervision signal based on the softmax scores of proximal depths. Additionally, we propose Confidence-Gated Dynamic Drafting to allow the model to adaptively extend the drafting depth, which maximizes accepted tokens while proactively truncating unreliable paths in high-entropy scenarios. In conclusion, our primary contributions can be listed as:

\begin{itemize}[leftmargin=*]
    \item We propose \textbf{Progressive Curriculum Loss}, a trajectory-aware gating mechanism that utilizes the cumulative product of preceding-depth probabilities to suppress gradient noise from unreliable distal predictions. This approach facilitates an automated easy-to-hard optimization transition, ensuring stable training even at extended look-ahead depths.
    \item We introduce \textbf{Confidence-Gated Dynamic Drafting} to maximize inference efficiency by adaptively calibrating speculative length based on cumulative joint probability. This mechanism allows the model to dynamically adjust its drafting reach to match the real-time prediction difficulty across varying document contexts.
    \item By integrating Progressive Curriculum Loss and Confidence-Gated Dynamic Drafting, our \mymodel\, achieves substantial acceleration (up to $5\times$) with negligible impact on parsing accuracy, marking the first successful validation of MTP with extensive look-ahead depth in the document parsing task.
\end{itemize}

\section{Related Work}

\subsection{VLM-based Efficient Document Parsing}
Recent advancements in document parsing have increasingly focused on optimizing inference performance to address high computational costs and latency. Existing approaches can be broadly categorized into two main directions based on their architectural paradigm: those employing a pipeline with a Vision-Language Model (VLM) as the decoder, and those conducting efficient adaptations directly upon a VLM backbone.

\noindent\textbf{Pipeline with VLM Decoder.} A prevalent strategy decomposes the complex parsing task into sequential, specialized stages. Representative methods~\cite{cui2025paddleocr,wang2024mineru} typically first employ a detector to locate distinct document elements (e.g., text blocks, tables), and then utilize a VLM decoder to recognize the content within each region serially. This paradigm allows for parallel processing across different regions, improving throughput for structured documents.

\noindent\textbf{Efficient Adaptation on Pre-trained VLM.} Another line of research seeks to enhance efficiency by modifying or extending a single VLM's inference process. This includes designing lightweight architecture variants~\cite{feng2025dolphin,lei2026unirec,feng2026dolphin2} to reduce computational footprint. Other works introduce specific mechanisms into the VLM:for instance, DeepSeek-OCR-2~\cite{wei2026deepseek} compresses the visual token sequence to shorten the input context; GLM-OCR~\cite{GLM-OCR} integrates the Multi-Token Prediction architecture, a technique also validated in large language models like Qwen3-Next~\cite{yang2025qwen3} and DeepSeek-V3~\cite{liu2024deepseek}; and Youtu-parsing~\cite{yin2026youtu} employs custom mask placeholders to enable parallel token inference.

\subsection{Speculative Decoding with Multi-Token Prediction}

To improve the inference efficiency of LLMs and VLMs, recent research has integrated auxiliary drafting structures directly into the target model. These methods can be categorized by their architectural topology and optimization strategies.

\noindent\textbf{Architectural Evolution}. Existing designs primarily follow two paradigms. Parallel drafting ( Medusa ~\cite{cai2024medusa}, MTP ~\cite{gloeckle2024better}) predicts multiple future tokens simultaneously from a single hidden state. While fast, it lacks the causal dependencies between candidate tokens. Serial drafting (Eagle~\cite{li2025eagle,li2024eagle,li2024eagle2},DeepSeek-V3 ~\cite{liu2024deepseek}), on the other hand, adopts an autoregressive structure at the feature level, capturing deterministic context more effectively. However, due to the challenges of training stability and error accumulation, existing methods in both categories typically employ only a limited number of heads (typically $ \le 3 $ ), which restricts the maximum theoretical speedup.

\noindent\textbf{Loss Function Optimization}. The training of these auxiliary heads follows two distinct loss weighting paradigms. Uniform loss (e.g., DeepSeek-V3 ~\cite{liu2024deepseek}) assigns equal weights to all prediction heads, emphasizing long-range dependencies but often suffering from high variance in distant token predictions. In contrast, position-dependent loss (Medusa  ~\cite{cai2024medusa}) utilizes a weighting scheme that decays as the head index increases. This approach prioritizes the convergence of immediate predictions to handle the escalating uncertainty of the far future. Despite these efforts, how to effectively scale the number of heads in serial architectures through loss-tuning remains an under-explored area, as simple weighting schemes often fail to balance the accuracy and quantity of serial drafting steps.

\section{Methodology}
\wu{Document parsing, which maps an image $\mathbf{I}$ to a text sequence $\mathbf{X} = \{x_1, \dots, x_T\}$, is a perceptual-heavy task where tokens are primarily driven by visual features. This heavy reliance on image evidence typically results in highly peaked predictive distributions, providing a reliable foundation for look-ahead decoding via MTP. In the following sections, we will revisit the mechanisms and challenges of MTP, followed by our proposed solution, P-MTP, for both training and inference.}

\subsection{\wu{Revisiting MTP}}
\xie{To begin with, let us review the inference acceleration mechanism of MTP and its theoretical speedup ratio.} \wu{Standard Next Token Prediction (NTP) employs an autoregressive decoding paradigm where a single forward pass generates only one token $\{x_t\}$, posing a substantial computational bottleneck for long document parsing. MTP addresses this efficiency constraint by predicting additional $K$ tokens $\{\hat{x}_{t+1}, \dots, \hat{x}_{t+K}\}$ simultaneously in one forward pass.}
As detailed in the Supplementary Material, the theoretical speedup ratio $\mathcal{S}$ of MTP can be formulated as:
\begin{equation}
\label{eq:speedup}
    \mathcal{S} \approx \frac{1 + \alpha K}{1 + \gamma K}
\end{equation}
where $\alpha \in [0, 1]$ is the average acceptance rate of \xie{the $K$ MTP} tokens, and $\gamma$ denotes the per-step overhead ratio of MTP Module. Evidently, maximizing the product $\alpha K$ is crucial for the speedup of MTP, especially when $\gamma$ is kept small through a lightweight MTP design. However, increasing $K$ to extend the prediction horizon often comes at the cost of a lower $\alpha$, primarily because long-range temporal dependencies are harder to capture accurately.
Furthermore, a static $K$ is suboptimal as it fails to dynamically maximize $\alpha K$ across varying contexts, either suffering from a collapsed $\alpha$ in complex scenarios or underutilizing the potential $K$ in simpler patterns.
Consequently, achieving efficient document parsing requires a transition from fixed-depth drafting to a more fluid mechanism. To this end, we introduce P-MTP, which utilizes Progressive Curriculum Loss to progressively scale look-ahead depth in the training phrase and Confidence-Gated Dynamic Drafting to enable adaptive inference acceleration based on real-time prediction difficulty.

\subsection{\wu{Training with Progressive Curriculum Loss}}
\begin{figure}[t]
    \centering
    \includegraphics[width=1\linewidth]{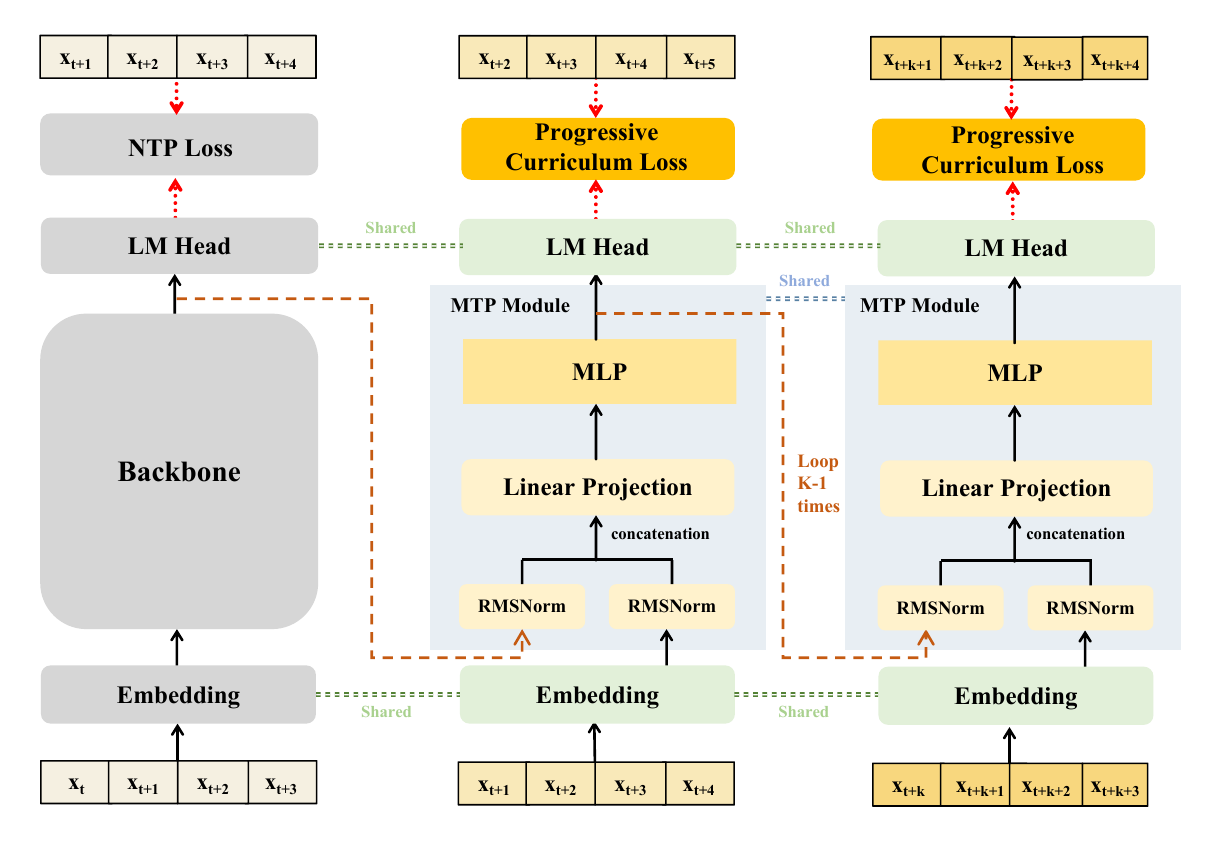}
    \caption{Architectural design for Learning P-MTP with Progressive Curriculum Loss. MTP Module is shared among different look-ahead depths. In the training phase, the framework conducts $K$ times recurrent look-ahead predictions with shared MTP Module, where the training difficulty increases as the look-ahead depth becomes deeper. By employing Progressive Curriculum Loss, the model gradually learns to capture long-range dependencies from $x_{t+2}$ to $x_{t+K+1}$.}
    \label{fig:mtp_module}
\end{figure}

\wu{Following prominent architectures~\cite{liu2024deepseek}, our tailored P-MTP for document parsing utilizes shared MTP Module to recurrently produce $K$ look-ahead predictions in a single forward pass. To mitigate the training instability inherent in distal token supervision, we formulate Progressive Curriculum Loss  that dynamically scales the supervision signal across the look-ahead depth, as illustrated in Figure~\ref{fig:mtp_module}.}

\noindent\textbf{MTP Module.} Building upon the primary hidden state $h_{t}^0$, MTP Module iteratively derives the $k$-th state $h_{t}^k$ by fusing the preceding state with the corresponding token embedding $e_{t+k}$:
\begin{equation}
h_{t}^k = \mathcal{F}_{\text{proj}}(\mathcal{F}_{\text{fuse}}(h_{t}^{k-1}, e_{t+k})), \quad k = {1, \dots, K}
\end{equation}
where $\mathcal{F}_{\text{fuse}}$ performs RMSNorm followed by concatenated linear reduction. For $\mathcal{F}_{\text{proj}}$ we adopt a lightweight residual MLP architecture maintaining sufficient representative power for the structural patterns inherent in document parsing. Finally, the shared LM Head $\mathcal{F}_{\text{cls}}$ is applied to yield the logits $\hat{p}_t^k = \mathcal{F}_{\text{cls}}(h_{t}^k)$.While prior serial architectures such as DeepSeek-V3 typically employ heavy transformer decoder layers for $\mathcal{F}_{\text{proj}}$ significantly increasing the per-step overhead ratio $\gamma$ (as defined in Equation ~\ref{eq:speedup}). Parallel classic approach like Medusa use independent heavy-weight heads which suffer from low acceptance rate $\alpha$ due to lacking non-linear representational capacity.

\begin{figure}[t]
    \centering
    \includegraphics[width=0.6\linewidth]{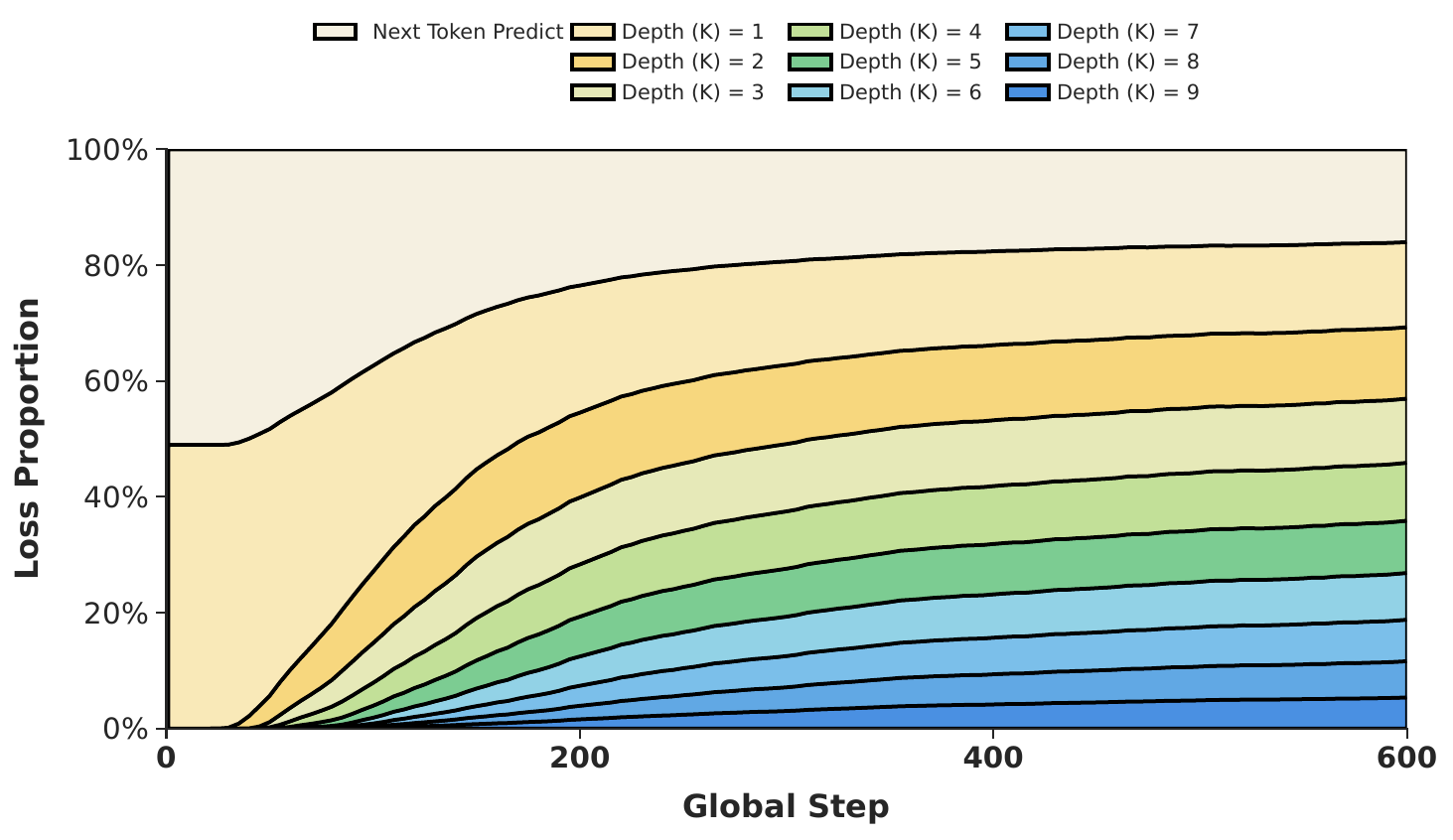}
    \caption{Evolution of loss proportion across training steps. The stacked area chart illustrates the progressive adjustment of training loss across varying look-ahead depths under Progressive Curriculum Loss. As training progresses, the loss distribution shifts monotonically toward deeper look-ahead depths, demonstrating an automatic transition from proximal to distal predictive tasks.}
    \label{fig:visualization_2}
\end{figure}

\noindent\textbf{Progressive Curriculum Loss.}
The total training objective $\mathcal{L}$ is typically defined as the joint optimization of the primary NTP and the look-ahead prediction tasks with MTP Module:
\begin{equation}
\label{eq:total}
\mathcal{L} = \mathcal{L}_{\text{NTP}} + \mathcal{L}_{\text{MTP}}.
\end{equation}
Specifically, $\mathcal{L}_{\text{NTP}}$ maintains the foundational language modeling capability by minimizing the negative log-likelihood over a sequence of length $T$:
\begin{equation}
\mathcal{L}_{\text{NTP}} = -\frac{1}{T} \sum_{t=0}^{T-1} \log p_{t}[x_{t+1}],
\end{equation}
where $p_t[x_{t+1}]$ represents the predicted probability by primary model at position $t$. Complementing the primary objective, $\mathcal{L}_{\text{MTP}}$ is formulated as a weighted sum of cross-entropy losses across $K$ look-ahead depths:
\begin{equation}
\label{eq:mtp}
\mathcal{L}_{\text{MTP}} = - \sum_{k=1}^{K} \omega_k \left(\frac{1}{T-k} \sum_{t=0}^{T-k-1} \log \hat{p}_{t}^k[x_{t+k+1}] \right).
\end{equation}
Herein, the hyperparameters $\{\omega_k\}_{k=1}^K$ determine the contribution of each look-ahead depth, while $\hat{p}_{t}^k[x_{t+k+1}]$ denotes the probability assigned to the ground-truth token at position $(t+k+1)$ by the shared head layer at the $k$-th look-ahead depth. By manually pre-defining weights $\omega_k$, prior methods heuristically suppress distal objectives to stabilize the training process. However, this approach struggles as the look-ahead depth $K$ scales, as distal objectives fail to be optimized significantly due to their vanishingly small yet fixed weights. To bridge this gap,
we introduce Progressive Curriculum Loss, which facilitates the optimization process by assigning a position-specific weight $\omega_{(t,k)}$ to each look-ahead depth $k$ at position $t$. This dynamic weighting scheme is decomposed into two complementary components:
\begin{itemize}
    \item \textbf{Sequential Path Constraint ($\omega^1$).} In a sequential prediction task starting at position $t$, the validity of prediction at look-ahead depth $k$ is strictly contingent upon the correctness of all its predecessors. We define sequential path weight $\omega^1$ as the cumulative product of the probabilities assigned to the ground-truth tokens at all previous depths:
    \begin{equation}
    \omega^{1}_{(t,k)} = \prod_{j=0}^{k-1} \hat{p}_{t}^{j}[x_{t+j+1}],
    \end{equation}
    where $\hat{p}_{t}^{0}[x_{t+1}]$ is defined as $p_t[x_{t+1}]$ which is predicted by primary model. Under this constraint, if the model yields a low-probability prediction at a proximal depth, the supervision signals for all distal tokens are exponentially suppressed.
    \item \textbf{Retrospective Target Constraint ($\omega^2$).} Parallel to the sequential path constraint, the retrospective target weight $\omega^2$ evaluates the historical consistency of predictions for the same target token $x_{t+k+1}$ across proximal look-ahead distances:
\begin{equation}
\omega^{2}_{(t,k)} = \prod_{j=0}^{k-1} \hat{p}_{t+j}^{k-j}[x_{t+k+1}].
\end{equation}
This weight ensures that the supervision for target $x_{t+k+1}$ at $k$-th look-ahead depth is prioritized only if the model shows stable convergence toward this specific position across the primary head and preceding look-ahead attempts. If the model previously struggled to `see' this target from closer distances, the current prediction is treated as unreliable noise and its loss is de-emphasized.
\end{itemize}
The final dynamic weight $\omega_{t,k}$ for the $k$-th look-ahead prediction at position $t$ is calculated as the product of these two factors, effectively gating the contribution of each look-ahead task based on both the reliability of the current trajectory and the predictive consistency of the target from proximal distances:
\begin{equation}
\label{eq:weight}
\omega_{(t,k)} = \omega^{1}_{(t,k)} \cdot \omega^{2}_{(t,k)}.
\end{equation}
Integrating this re-weighting strategy, the dynamic MTP loss $\mathcal{L}_{\text{P-MTP}}$ is formulated as a dynamically weighted sum of cross-entropy losses:
\begin{equation}
\mathcal{L}_{\text{P-MTP}} = - \sum_{k=1}^{K} \left( \frac{1}{T-k} \sum_{t=0}^{T-k-1} \omega_{(t,k)} \cdot {\text{log}}(\hat{p}_{t}^k[x_{t+k+1}]) \right). \end{equation}
By substituting $\mathcal{L}_{\text{MTP}}$ with $\mathcal{L}_{\text{P-MTP}}$ in Equation ~\ref{eq:total}, our model effectively implements progressive curriculum learning for MTP learning. Unlike conventional MTP approaches that assign static, pre-defined weights to all look-ahead depths from the outset, our Progressive Curriculum Loss inherently prioritizes simpler, high-confidence prediction paths during the early stages of training. This stabilizing mechanism is evidenced in Figure ~\ref{fig:visualization_2}, where the loss proportion for deeper look-ahead depths increase only after shorter-term trajectories have stabilized, ensuring a robust easy-to-hard curriculum learning.

\subsection{\wu{Inference via Confidence-Gated Dynamic Drafting}}
Although deeper look-ahead has been explored during training, employing a fixed drafting length $K$ in inference often fails to fully exploit the acceleration potential of MTP. While simple text blocks allow for deep look-ahead, complex semantic transitions require cautious step-by-step generation, making a static draft length inherently inefficient. To address this, we propose Confidence-Gated Dynamic Drafting to adaptively determine whether to continue or terminate drafting based on real-time prediction confidence.

\begin{figure}[t]
    \centering
    \includegraphics[width=1\linewidth]{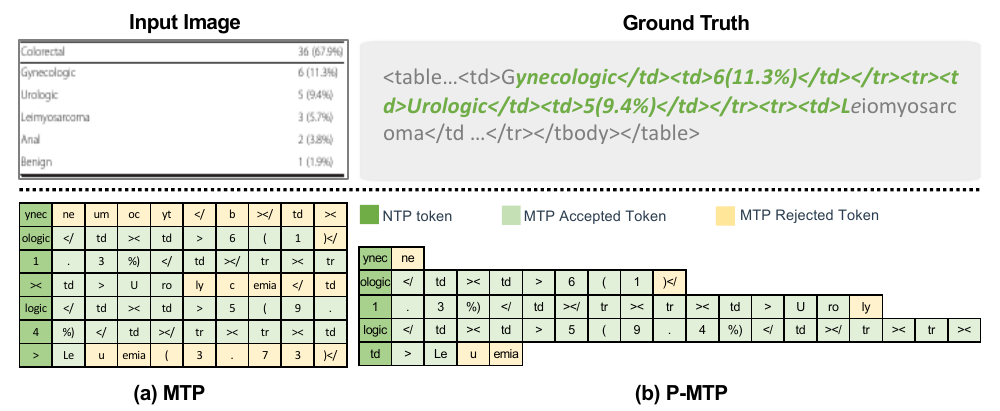}
    \caption{Comparison between standard MTP and the proposed \mymodel\, with Confidence-Gated Dynamic Drafting during inference. Unlike the fixed look-ahead depth used in (a) MTP, (b) \mymodel\, dynamically adjusts the drafting length to match inference difficulty. Such flexibility enables the model to reach an optimal $\alpha \times K$ depth when confidence is high and prune sequences when certainty is low, significantly reducing rejected tokens and computational waste.}
    \label{fig:visualization_3}
\end{figure}
\noindent\textbf{Confidence-Gated Dynamic Drafting.} Unlike vanilla speculative decoding that employs a fixed drafting length, \mymodel\, treats the look-ahead depth as a stochastic variable constrained by the cumulative joint probability of the predicted sequence. Although the model is trained with a look-ahead depth $K$, empirical observations suggest that it possesses the predictive capacity to generalize to extended contexts during inference. This also  attributed to our serial architecture and weight sharing across multiple depths, which enhance generalization and coherence.We thus employ an expanded budget $H = 2K$ and dynamically determine the optimal draft depth $k^*$, which can be formulated as:
\begin{equation}
k^* = \max \{ k \in \{1, \dots, H\} \mid \prod_{j=0}^{k-1}\hat{p}_{t}^{j}[\hat{x}_{t+j+1}] > \delta \},
\end{equation}
where $\delta \in [0, 1]$ is a predefined confidence threshold based on final training status. This formulation treats the product of probabilities as a real-time proxy for the joint reliability in the inference phrase, allowing \mymodel \ to adaptively exhaust the look-ahead potential when the content is deterministic and halt early when uncertainty arises. A comparative visualization is provided in Figure \ref{fig:visualization_3}. Compared to vanilla MTP (a), which suffers from a low acceptance ratio due to its fixed look-ahead depth, \mymodel\, (b) effectively concentrates the speculative tokens on highly probable trajectories, thereby maximizing the effective throughput of each decoding step.

\noindent\textbf{Reliability-Aware $\delta$ Calibration.} We propose a calibration strategy for $\delta$ that explicitly maps inference sensitivity to \mymodel  training dynamics. Given that distal token supervision is exponentially suppressed by dual constraints $\omega^1$ and $\omega^2$, we define $\delta$ as a function of the terminal empirical loss $\bar{\mathcal{L}}$ and the look-ahead depth $K$:
\begin{equation}
\delta = \delta_{\text{base}} \cdot e^{\frac{\lambda \cdot  \bar{\mathcal{L}}}{K}},
\end{equation}
where $\bar{\mathcal{L}}$ denotes the mean empirical loss evaluated on the validation set upon training convergence, $K$ represents the total look-ahead depths in training, and $\lambda$ is a sensitivity coefficient typically assigned a value of 2 to counteract the effects of dual-path suppression. The parameter $\delta_{\text{base}}$ serves as the empirical baseline threshold, initialized to 0.3.
Unlike heuristic-based thresholding, reliability-aware $\delta$ scales positively with the residual uncertainty $\bar{\mathcal{L}}$. For a model with higher convergence loss, $\delta$ automatically shifts towards a conservative regime to filter out hallucinated distal tokens. Conversely, as $\bar{\mathcal{L}}$ decreases, the threshold relaxes to $\delta_{base}$, facilitating more aggressive look-ahead. This calibration minimizes the discrepancy between the training-time curriculum and inference-time execution, ensuring robust speculative performance across diverse semantic complexities.

\section{Experiments}

\subsection{Experimental Setup}
\noindent\textbf{Benchmarks.} To evaluate the efficacy of the proposed \mymodel\, based document parsing framework, we conduct extensive experiments across three representative tasks: formula recognition, table structure recognition, and general document parsing on \wu{UniMERNet}, \wu{PubTabNet}, and \wu{OmniDocBench} respectively. These datasets cover a wide range of document layouts and structural complexities, providing a robust testbed for both parsing accuracy and generation efficiency.

\noindent\textbf{Evaluation Protocol.} We adopt the Complexity-aware Document Metric (CDM) as the primary evaluation criterion~\cite{wang2024cdm}. CDM accounts for the structural nesting and semantic significance of LaTeX tokens, providing a more robust and fair assessment of mathematical expressions compared to simple string-based metrics. TEDS score~\cite{zhong2020image} is adopted to measure the similarity between the predicted table and the ground truth, effectively capturing the structural alignment and content accuracy of the parsed table. For general document parsing, we report the overall performance using the offical evaluation protocol following previous works~\cite{ouyang2025omnidocbench}.

\noindent\textbf{Implementation details.} We adopt InternVL3.5-1B and Qwen3-VL-2B  trained with the standard  Supervised Fine-tuning (SFT) as baselines to demonstrate the versatility of our approach. For formula and table recognition, we utilize the official training sets corresponding to each benchmark. For general document parsing, the models are trained on \wu{public dataset from LightOnOCR-2~\cite{taghadouini2026lightonocr} consisting of 0.42M samples}. The training consists of two epochs: the first epoch follows the standard supervised fine-tuning paradigm to establish a strong foundational parsing capability. In the second epoch, we integrate the MTP objective by performing supervision of Equation ~\ref{eq:total}. We employ the Adam optimizer~\cite{dettmers20218} with layer-wise learning rate scheduling: a peak learning rate of \wu{$1e-5$} for the backbone and \wu{$5e-5$} for the MTP module, with a warmup ratio of \wu{0.05}. The global batch size is set to \wu{8}. During the inference phase, we first evaluate our model using the vanilla model forward to measure the most direct acceleration ratio. Furthermore, we integrate \mymodel, into the VLLM framework to achieve a highly optimized version for practical deployment. All performance benchmarks are conducted on a single NVIDIA A100 40GB GPU.

\subsection{Effectiveness of \mymodel}
\begin{figure}[h]
    \centering
    \includegraphics[width=0.8\linewidth]{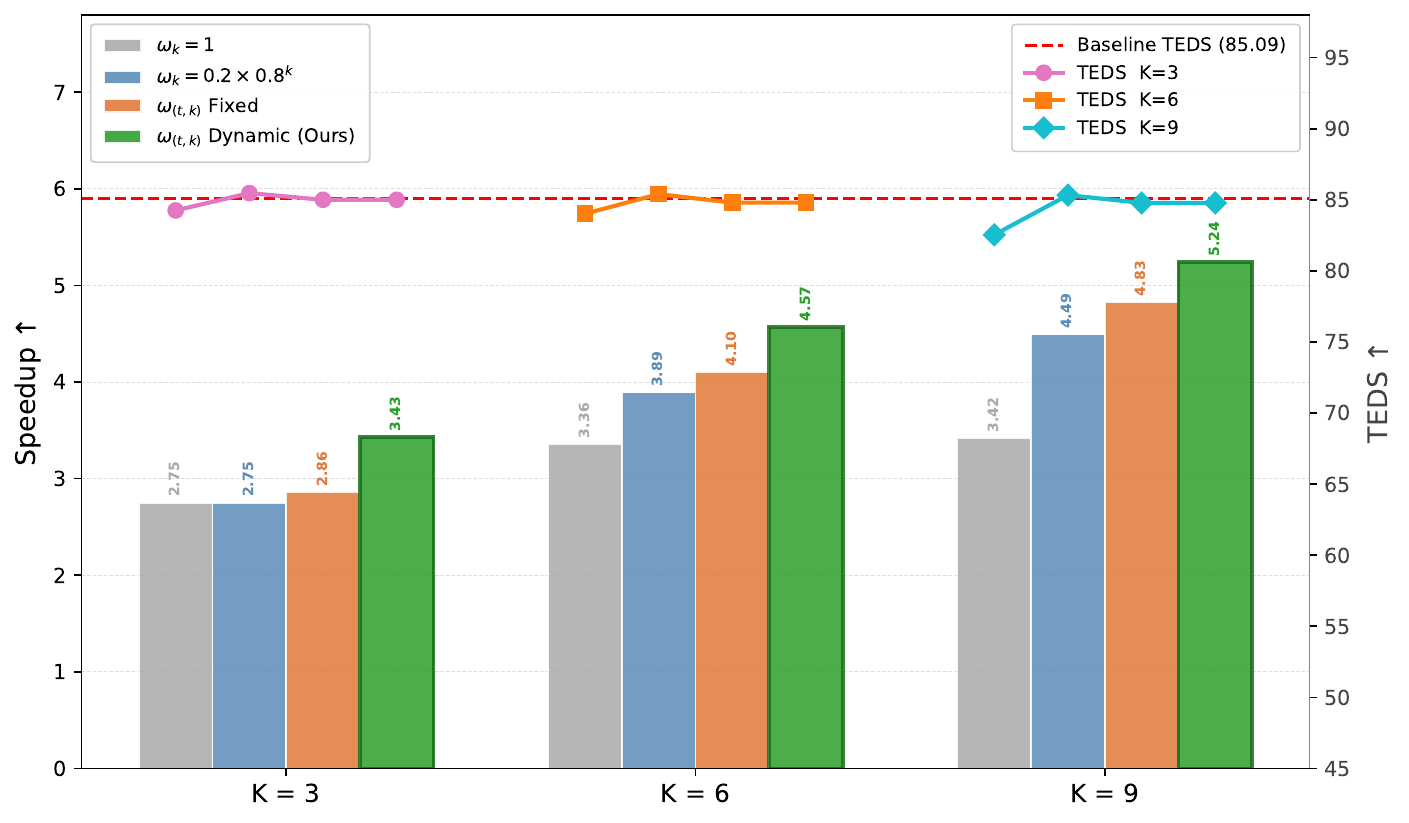}
    \caption{Speedup and TEDS performance under different values of $K$ and four distinct weight schemes.}
    \label{mtp_accept_length}
 \end{figure}

To verify the individual contributions and synergistic effects of our proposed components, we conducted extensive ablation studies on the PubTabNet dataset specifically for the table recognition task. All ablation experiments were implemented using Qwen3-VL-2B as the base model. As shown in Figure~\ref{mtp_accept_length}, after applying \text{P-MTP}, the model maintains accuracy comparable to the baseline under various progressively increasing look-ahead depth, while achieving a maximum speedup ratio of 5.24 $\times$.

\noindent\textbf{Ablation Studies on Alternative MTP architectures.} We conduct a comprehensive ablation study to evaluate the impact of different architectural choices within the MTP module, as summarized in Table~\ref{tab:ablation_architectures}. A simple parallel LM head ~\cite{cai2024medusa} performs decently but is outperformed by more parameterized designs such as MLP or DecoderLayer in accept length. While the serial DecoderLayer ~\cite{wei2026deepseek} offers strong modeling capacity, it introduces significant latency overhead. Serial MLP achieves the best overall performance, while maintaining a high TEDS score and achieving the lowest latency. It strikes an optimal balance, providing sufficient non-linear modeling capability to predict multiple future tokens accurately without becoming a computational bottleneck.

\begin{table}[h]
\centering
\setlength{\tabcolsep}{5pt} 
\renewcommand{\arraystretch}{1.2} 
\caption{Comparison of MTP-related configurations with varying architectures, head strategies, and projection layers $\mathcal{F}_{\text{proj}}$. $\tau$ denotes the average acceptance length.}
\label{tab:ablation_architectures}
\begin{tabular}{lllcccc}
\toprule
{Architecture} & LM Head & $\mathcal{F}_{\text{proj}}$  & {TEDS}$\uparrow$ & {Latency}$\downarrow$ & ${\tau}$$\uparrow$ \\
\midrule
Parallel    & Separate &    -  & 83.98 & 5.85 & 6.28  \\
Parallel  &  Shared  & MLP   & 84.44 & 5.62 &7.19  \\
Parallel  & Shared   & DecoderLayer & 84.09 & 9.53 & 7.26 \\
Serial  & Shared   & DecoderLayer & \textbf{84.84} & 7.85 & 6.27  \\
Serial &   Shared & MLP  & 84.78 &\textbf{5.53}  & \textbf{7.29}  \\
\bottomrule
\end{tabular}
\end{table}

\begin{table}[htb]
\centering
\setlength{\tabcolsep}{3pt} 
\renewcommand{\arraystretch}{1.1} 
\caption{Ablation study on loss weighting schemes and drafting configurations. We compare our proposed dynamic loss weight $\omega_{(t,k)}$ against standard $\omega_{k}$ on PubTabNet. `Fixed' refers to vanilla speculative decoding with a fixed drafting length, while `Dynamic' represents inference with our proposed Confidence-Gated Dynamic Drafting.$\tau$ denotes the average acceptance length.}
\label{tab:ablation_weight}
\begin{tabular}{llcccc}
\toprule
{{\makecell[l]{Weighting\\Scheme}}} & {{\makecell[l]{Drafting\\Setting}}} \quad\quad & {TEDS}$\uparrow$ & {Latency}$\downarrow$ & ${\tau}$$\uparrow$ & {Speedup}$\uparrow$ \\
\midrule
\rowcolor{shadecolor}
\multicolumn{6}{c}{{Base (NTP)}} \\
{-} & - &  85.09 & 26.71 & 1 & 1 \\
\midrule
\rowcolor{shadecolor}
\multicolumn{6}{c}{\textit{$K=3$}} \\
$\omega_{k}=1$ & Fixed & 84.25 & 9.72  & 3.79 & 2.75 \\
$\omega_{k} = 0.2\times 0.8^k$ & Fixed & 85.47 & 9.72 & 3.77 & 2.75 \\
$\omega_{(t,k)}$  & Fixed  & 85.00 &  9.34  & 3.81 & 2.86 \\
$\omega_{(t,k)}$  &  Dynamic & {85.00} &  \textbf{7.79}  & \textbf{5.39}& \textbf{3.43} \\
\hline
\rowcolor{shadecolor}
\multicolumn{6}{c}{\textit{$K=6$}} \\
$\omega_{k} = 1$ & Fixed &84.02 & 7.95 & 5.60& 3.36 \\
$\omega_{k} = 0.2\times 0.8^k$ & Fixed & 85.40 & 6.86 & 5.82 & 3.89\\
$\omega_{(t,k)}$ & Fixed & 84.80 & 6.52  & 5.93& 4.10 \\
$\omega_{(t,k)}$  &  Dynamic  & {84.80} & \textbf{5.85}  & \textbf{7.67}& \textbf{4.57} \\

\hline
\rowcolor{shadecolor}
\multicolumn{6}{c}{\textit{$K=9$}} \\
$\omega_k = 1$ & Fixed & 82.54 & 7.82 & 5.85 & 3.42 \\
$\omega_{k} = 0.2\times 0.8^k$  & Fixed & 85.33 & 5.95  & 7.05 & 4.49 \\
$\omega_{(t,k)}$ & Fixed & 84.78 &5.53 & 7.29 & 4.83 \\
$\omega_{(t,k)}$  & Dynamic & {84.78} &\textbf{5.10} & \textbf{8.60} & \textbf{5.24} \\

\bottomrule
\end{tabular}
\end{table}

\noindent\textbf{Effectiveness of Progressive Curriculum Loss.} To evaluate the efficacy of the proposed Progressive Curriculum Loss, the dynamic weighting strategy $\omega_{(t,k)}$ was compared against two representative baseline schemes: an equal weighting strategy ($\omega_{k}=1$), commonly adopted by models such as DeepSeek-V3, and a static decay weighting strategy ($\omega_{k}=0.2 \times 0.8^k$), a heuristic approach utilized in frameworks like Medusa. As can be easily observed from Table \ref{tab:ablation_weight}, the proposed loss weighting mechanism demonstrates robust scalability by effectively extending the look-ahead depth to $K=9$, a significant improvement over the typical depth of $K \leq 3$ observed in most prior works. Specifically, even at a look-ahead depth of $K=9$, the proposed method maintains a TEDS score of 84.78, which is highly comparable to the Base (NTP) performance, while simultaneously achieving a substantial speedup of 4.83$\times$. Furthermore, a comparative analysis against alternative loss weighting schemes reveals that the dynamic weighting strategy $\omega_{(t,k)}$ consistently achieves best speedup across all tested drafting lengths compared with existing approaches.

To further evaluate the components within $\omega_{(t,k)}$, we conducted an ablation study on the sequential constraint ($\omega^1$) and retrospective constraint ($\omega^2$). Table ~\ref{tab:ablation_loss_merge} shows that while each constraint independently reaches attain a significant speedup, their synergistic integration ($\omega^1 \times \omega^2$) yields the best results, reaching a peak speedup of 4.83$\times$. This confirms that $\omega^1$ and $\omega^2$ collectively drive the efficacy of our proposed \mymodel.

\begin{table}[h]
\centering
\setlength{\tabcolsep}{8pt} 
\renewcommand{\arraystretch}{1.2} 
\caption{Ablation study on weighting mechanisms for $\omega_{(t,k)}$. By dynamically combining sequential and retrospective constraints (Eq. ~\ref{eq:weight}), our method achieves a superior trade-off between TEDS and efficiency.}
\label{tab:ablation_loss_merge}
\begin{tabular}{lcccc}
\toprule
{Method} & {TEDS}$\uparrow$ &  {Latency}$\downarrow$  & $\boldsymbol{\tau}$$\uparrow$ &  {Speedup}$\uparrow$\\
\hline
Base & 85.09 & 26.71 & 1 & 1 \\

$\omega^1$& 84.16 & 6.33 & 6.95 & 4.22\\
$\omega^2$& 84.37 & 6.09 & 7.06 & 4.39\\
$\omega^1 \times \omega^2$& \textbf{84.78} & \textbf{5.53}  & \textbf{7.29 } & \textbf{4.83}\\
\bottomrule
\end{tabular}
\end{table}

\noindent\textbf{Effectiveness of Confidence-Gated Dynamic Drafting.} A comparative analysis across different drafting length reveals the robust scalability of our Confidence-Gated Dynamic Drafting. As shown in Table ~\ref{tab:ablation_weight}, the dynamic configuration consistently yields superior efficiency gains over the fixed-length baseline across varying training look-ahead depth $K$. Specifically, for $K=3, 6,$ and $9$, the dynamic approach improves the speedup from 2.86$\times$ to 3.43$\times$, 4.10$\times$ to 4.57$\times$, and 4.83$\times$ to 5.24$\times$, respectively. This consistent upward trajectory in speedup is directly coupled with a substantial increase in the average acceptance length ($\tau$), which sees absolute gains of $+1.58$, $+1.74$, and $+1.31$ for each respective $K$. The results  provide strong empirical evidence that dynamically adjusting drafting length based on real-time inference difficulty effectively minimizes redundant computations and maximizes decoding efficiency.

\begin{table}[htbp]
\centering
\small
\setlength{\tabcolsep}{8pt} 
\renewcommand{\arraystretch}{1.2} 
\caption{Throughput(TPS, tokens per second) speedup of our proposed P-MTP method relative to the baseline (without speculative sampling) across different batch sizes in vLLM framework.}
\label{tab:ablation_vllm}
\begin{tabular}{lcccc}
\toprule
\textbf{Batch Size} & \textbf{4} & \textbf{16} & \textbf{32} & \textbf{64}  \\
\hline
Qwen3VL-2B$_{\text{Base}}$ & 647 & 1894 & 2774 & 3719\\
Qwen3VL-2B$^{\dagger}_{\text{P-MTP}}$ & 1028 & 2921 &  4315 & 5638\\
\midrule
\textbf{Speedup} & \textbf{1.59$\times$} & \textbf{1.54$\times$} & \textbf{1.56$\times$} & \textbf{1.52$\times$}\\
\bottomrule
\end{tabular}
\end{table}
\noindent\textbf{Throughput Evaluation.}To better adapt our method to real-world production environments, we further implemented P-MTP based on the vLLM \cite{kwon2023efficient}, a popular high-performance inference framework for production. $\text{Method}_{\text{Base}}$ denotes the model trained with the standard SFT and executed via vanilla NTP inference. In contrast, $\text{Method}^{\dagger}_{\text{P-MTP}}$ employs our proposed Progressive Curriculum Loss with a look-ahead depth of 9 during training and utilizes speculative decoding powered by Confidence-gated Dynamic Drafting during inference. The corresponding results under different batch sizes are summarized in Table~\ref{tab:ablation_vllm}, achieving a speedup of 1.5$\times$.

\subsection{Generality across Parsing Frameworks.}

We also validate the benefits of the \text{P-MTP} across different base models and tasks as shown in Table ~\ref{tab:main_results}. As can be easily seen, the integration of the proposed \mymodel\, across different base models ( InternVL3.5-1B and Qwen3-VL-2B) shows significant speedup while maintaining high accuracy. Specifically, the \text{P-MTP} variants of different models achieve 7.48-9.66 accept length on the Formula task, 8.48-8.60 on the Table task, and 3.05-3.45 on the Document task compared with the Base model, with slight fluctuations in precision. Different accept lengths resulted in a throughput acceleration of $1.4\times$-$2.3\times$. This confirms that P-MTP is not only theoretically sound but also exceptionally effective when deployed within industry-standard high-performance inference frameworks across different tasks and base models.

 \begin{table}[h]
   \centering
  \caption{Performance and inference speed comparison across diverse document parsing tasks. We report task-specific metrics (CDM for Formula, TEDS for Table, Overall for Document), TPS(tokens per second), and average acceptance length.} 
  \label{tab:main_results}
  \resizebox{\textwidth}{!}{\begin{tabular}{l|ccc|ccc|ccc}
  
   \toprule
    \multirow{2}{*}{Method} & \multicolumn{3}{c|}{\textbf{Formula}} &  \multicolumn{3}{c|}{\textbf{Table}}&  \multicolumn{3}{c}{\textbf{Document}} \\
    \cmidrule(lr){2-10}
     & CDM$\uparrow$ & TPS$\uparrow$ &  $\boldsymbol{\tau}\uparrow$  & TEDS$\uparrow$ & TPS$\uparrow$ &  $\boldsymbol{\tau}\uparrow$ & Overall$\uparrow$ & TPS$\uparrow$ &  $\boldsymbol{\tau}\uparrow$ \\
 
  \hline
 \addlinespace
  InternVL3.5-1B$_{\text{Base}}$ & 95.64 & 1136 & 1       & 83.40 & 2475 & 1         & 73.30 & 1324 & 1 \\
  InternVL3.5-1B$^{\dagger}_{\text{P-MTP}}$ &  92.77 & 1684 & 7.48         & 86.26 & 5816 & 8.48        & 71.34 & 1822 & 3.05  \\
  
  \hline
  \addlinespace
  Qwen3VL-2B$_{\text{Base}}$ & 95.88 & 1649 & 1    & 85.09 & 2774 & 1        & 86.31 & 582 & 1  \\
  Qwen3VL-2B$^{\dagger}_{\text{P-MTP}}$ & 94.58 & 2663 & 9.66          & 84.78 & 4315 & 8.60       & 81.28 & 891 & 3.45  \\
  

  \bottomrule
 
\end{tabular}}
\end{table}

\section{Conclusion}
In this paper, we addresses the inherent efficiency bottlenecks in VLM-based document parsing. We identify that traditional MTP approaches suffer from static loss re-weighting, which restricts the model to shallow look-ahead depths and hinders the delicate trade-off between predictive performance and decoding speed. We proposed \mymodel, a trajectory-aware framework that scales look-ahead horizons through Progressive Curriculum Loss and Confidence-Gated Dynamic Drafting. Our extensive validation across different base models underscores the exceptional versatility and robustness of the \mymodel\, framework. In all scenarios, the proposed method achieved pronounced speedups with slight fluctuations in accuracy, bridging the gap between theoretical multi-token objectives and practical deployment requirements. These findings mark a breakthrough in document parsing, proving that \mymodel\, is a robust and scalable solution for real-world document intelligence.

\bibliography{main}

\setcounter{figure}{0}
\makeatletter
\renewcommand{\thefigure}{A\@arabic\c@figure}
\makeatother

\setcounter{table}{0}
\makeatletter
\renewcommand{\thetable}{A\@arabic\c@table}
\makeatother

\clearpage
\newpage
\appendix

\section*{Appendix}
\label{sec:appendix}

\section{Additional Results on Model Scaling Behavior}
\label{sec:app_model_scaling}
In this section, we further investigate the scaling behavior of our proposed framework across different model capacities (2B, 4B, and 8B parameters). As detailed in Table \ref{tab:ablation_model_size}, all variants are based on the Qwen3-VL-Instruct architecture trained on the PubTabNet dataset, and we compare our method against the standard autoregressive baselines.
\begin{table}[htbp]
\centering
\setlength{\tabcolsep}{10pt} 
\renewcommand{\arraystretch}{1.2} 
\caption{Full results for model size scaling.}
\label{tab:ablation_model_size}
\begin{tabular}{clccc}
\toprule
Model Size & Method & TEDS$\uparrow$ & Latency$\downarrow$ & $\tau$$\uparrow$ \\
\midrule
\multirow{2}{*}{2B} 
& Baseline & 85.09 & 26.71 & 1.00 \\
&  \textbf{$+$P-MTP (Ours)}  & 84.78 & 5.10  & 8.60 \\
\midrule
\multirow{2}{*}{4B} 
& Baseline & 86.68 & 34.25     & 1.00 \\
& \textbf{$+$P-MTP (Ours)}  & 86.60 & 8.43  & 10.92 \\
\midrule
\multirow{2}{*}{8B} 
& Baseline & 86.30 & 45.14 & 1.00 \\
& \textbf{$+$P-MTP (Ours)}   & 86.20 & 10.21 & 11.67 \\
\bottomrule
\end{tabular}
\end{table}

  
 
  
  
 
An important observation from the empirical results is that the drafting efficiency of our method exhibits a positive scaling trend with the size of the base model. Specifically, the average acceptance length $\tau$ increases monotonically with the model parameter count: from 8.60 for the 2B model, to 10.92 for the 4B model, and further to 11.67 for the 8B model. This positive correlation indicates that models with larger parameter sizes possess a superior capability to capture long-range dependencies and predict future contextual information, thereby generating longer and more accurate draft sequences during inference. Consequently, our framework achieves substantial latency reductions, with approximate speedups of 5.2$\times$, 4.1$\times$, and 4.4$\times$ for the 2B, 4B, and 8B variants, respectively. Crucially, this acceleration is achieved with negligible impact on downstream task performance. Across all scales, our method maintains a TEDS metric highly competitive with the baselines. In summary, these scaling trends demonstrate that our method is highly extensible and achieves significant acceleration effects when deployed across multimodal models of varying scales.

\section{Additional Results on Different Model}
\label{sec:app_diff_model}
To further evaluate the generalizability of our proposed framework, we extend our experiments beyond general-purpose multimodal foundational models to a specialized, lightweight architecture. Specifically, we integrate our P-MTP module into PaddleOCR-VL-1.5-0.9B, a compact model highly optimized for fine-grained document parsing tasks.

As summarized in Table \ref{tab:additional_diff_model} , our method consistently delivers efficiency improvements without compromising generation quality. For the table parsing task, the P-MTP module achieves a remarkable average acceptance length ($\tau$) of 6.64. This indicates that our drafting strategy remains highly effective at capturing structural dependencies in the specialized VLM. Consequently, this robust drafting capability yields a substantial inference acceleration, elevating the tokens per second (TPS) from 4454 to 6034 (an approximate $1.35\times$ speedup). Crucially, this acceleration is achieved while strictly preserving and marginally improving task performance. The TEDS metric for table parsing increases slightly from 81.05 to 82.03, confirming that our method reliably ensures structural fidelity during the accelerated generation process.

Overall, these supplementary results strongly demonstrate that our proposed module is model-agnostic. It can be seamlessly deployed on domain-specific, specialized architectures to provide substantial inference acceleration while rigorously maintaining accuracy.


\begin{table}[h]
   \centering
  \caption{Performance and inference speed comparison on P-MTP with PaddleOCR-VL-1.5.}
  \label{tab:additional_diff_model}
    \begin{tabular}{l|ccc}
      \toprule
      \multirow{2}{*}{Method} & \multicolumn{3}{c}{\textbf{Table}} \\
      \cmidrule(lr){2-4}
      & TEDS$\uparrow$ & TPS$\uparrow$ & $\boldsymbol{\tau}\uparrow$ \\
      \hline
      \addlinespace
      PaddleOCR-VL-1.5$_{\text{Base}}$ & 81.05 & 4454 & 1 \\
      PaddleOCR-VL-1.5$^{\dagger}_{\text{P-MTP}}$ & 82.03 & 6034 & 6.64 \\
      \bottomrule
    \end{tabular}
\end{table}

\section{Additional Results on Different Dataset}
\label{sec:app_diff_dataset}

We further evaluate our approach on a comprehensive document parsing task using Qwen3-VL-2B. In the main text, we adopt the open-source LightOnOCR-2 as the document parsing dataset to train both the SFT baseline and our P-MTP model, with the goal of validating the effectiveness of P-MTP on document-oriented tasks. However, the results reveal a non-trivial degradation in the overall score compared to the baseline. We attribute this degradation primarily to the limited scale and suboptimal quality of LightOnOCR-2, rather than to fundamental limitation of our method. Given the absence of publicly available document datasets that achieve strong performance on OmniDocBench, we resort to our in-house document dataset (approximately 1.5M samples) as the training dataset. As shown in Table~\ref{tab:inhouse_pageread_data_exp}, training with this high-quality in-house data yields only marginal fluctuation in the overall score relative to the baseline, while delivering a 2$\times$ improvement in inference throughput.

The results demonstrate that, when trained on sufficiently high-quality data, our P-MTP architecture can substantially accelerate inference with negligible impact on task performance.
\begin{table}[h]
   \centering
   \caption{Performance and inference speed comparison on document parsing . We report Overall metric, TPS(tokens per second), and average acceptance length $\boldsymbol{\tau}$.} 
   \label{tab:inhouse_pageread_data_exp}
   \begin{tabular}{l|ccc}   
  
   \toprule
    \multirow{2}{*}{Method} & \multicolumn{3}{c}{\textbf{Document}} \\
    \cmidrule(lr){2-4}
     & Overall$\uparrow$ & TPS$\uparrow$ &  $\boldsymbol{\tau}\uparrow$ \\
 
  \hline
  \addlinespace
  Qwen3VL-2B$_{\text{Base}}$ & 88.71 & 716 & 1  \\
  Qwen3VL-2B$^{\dagger}_{\text{P-MTP}}$ & 87.19 & 1437 & 7.20  \\
  
  \bottomrule
 
\end{tabular}
\end{table}

\section{Theoretical Justification for Threshold Calibration}
\label{sec:app_threshold}

To establish a principled mapping from the training objective to the inference-time threshold, we analyze the signal-to-noise ratio inherent in the P-MTP framework.

\noindent\textbf{Expected Supervision Signal Strength.} In P-MTP, the supervision signal for the $k$-th look-ahead head at position $t$ is gated by the dynamic weight $\omega_{(t,k)}$. This weight is the product of sequential path reliability $\omega^1$ and retrospective consistency $\omega^2$. Assuming an average prediction probability $\bar{p}$ for each token in a well-calibrated model, where $\bar{p} \approx e^{-\frac{\bar{\mathcal{L}}}{K}}$ (derived from the cross-entropy relationship $\mathcal{L} \approx -\ln p$), we can approximate the expected weights as:
\begin{itemize}
 \item Sequential Path Constraint: $\mathbb{E}[\omega^1_{(t,k)}] \approx \bar{p}^k = e^{-\frac{k \bar{\mathcal{L}}}{K}}$.
 \item Retrospective Target Constraint: $\mathbb{E}[\omega^2_{(t,k)}] \approx \bar{p}^k = e^{-\frac{k \bar{\mathcal{L}}}{K}}$.
\end{itemize}
Consequently, the joint supervision signal strength $\Omega$ for the $k$-th head during training follows a dual-path power law:
\begin{equation}
\mathbb{E}[\omega_{(t,k)}] = \mathbb{E}[\omega^1 \cdot \omega^2] \approx \bar{p}^{2k} = e^{-\frac{2k \bar{\mathcal{L}}}{K}}.
\end{equation}

\noindent\textbf{From Training Gating to Inference Drafting.} During inference, our Confidence Gated Dynamic Drafting employs a cumulative probability product $\prod_{j=0}^{k-1} \hat{p}^j$ to decide whether to continue the trajectory. This product is functionally equivalent to the sequential path weight $\omega^1_{(t,k)}$ used during training. However, a fundamental discrepancy exists: the $k$-th head was only effectively optimized during training when the signal $\omega_{(t,k)}$ was sufficiently high. To ensure that the drafted tokens at inference time maintain the same level of reliability consistency as the training supervision, the inference threshold $\delta$ must compensate for the "missing" retrospective constraint $\omega^2$ and the residual uncertainty $\bar{\mathcal{L}}$.

\noindent\textbf{Reliability Compensation Mapping.} While training utilizes the joint weight $\omega$, inference relies solely on the sequential product $\prod p$. To compensate for the missing retrospective constraint $\omega^2$ and the model's residual uncertainty $\bar{\mathcal{L}}$, the threshold $\delta$ must scale with the inverse of the expected training signal: $\delta \propto 1/\mathbb{E}[\omega_{(t,k)}]$.

To ensure $\delta$ serves as a robust recursive stopping criterion, we abstract the depth-specific decay into a global uncertainty density $\bar{\mathcal{L}}/K$. This ensures that $\delta$ acts as a $k$-invariant reliability gate that adaptively compensates for the terminal convergence state:
\begin{equation}
\delta = \delta_{\text{base}} \cdot e^{\frac{\lambda \cdot \bar{\mathcal{L}}}{K}},
\end{equation}
where $\lambda=2$ accounts for the dual-path suppression during training, and $\delta_{\text{base}}$ is the baseline confidence for an oracle model ($\bar{\mathcal{L}} \to 0$).

\begin{figure}[t]
    \centering
    \includegraphics[width=0.75\linewidth]{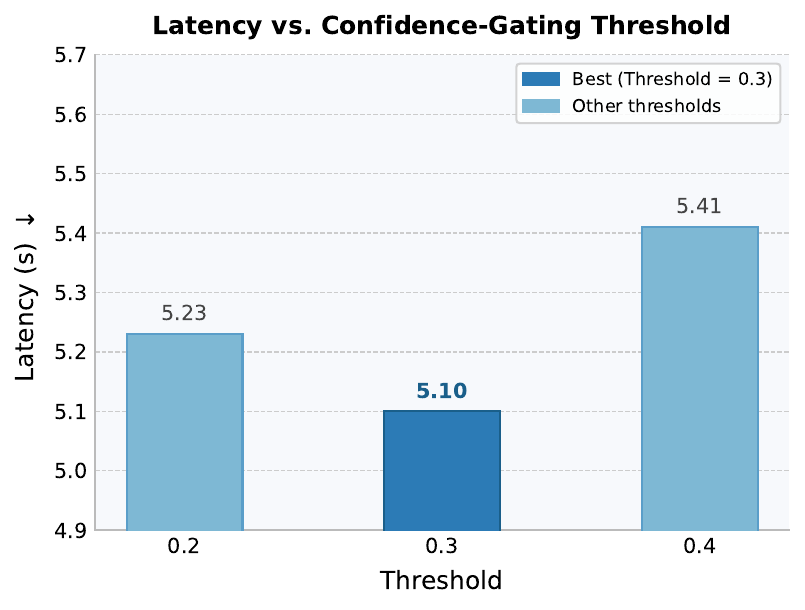}
    \caption{Latency comparison across different confidence thresholds using the Confidence-Gated Dynamic Drafting strategy. The threshold of 0.3 achieves the lowest latency.}
    \label{fig:threshold_latency}
\end{figure}

\noindent\textbf{Statistical Optimality of the Baseline Threshold.}
In a high-confidence, zero-loss scenario, we model the drafting process as a traversal over a low-entropy manifold where the expected joint probability follows $\mathbb{E}[\prod_{j=0}^{k-1} p_j] \approx \bar{p}^k$. $\delta_{\text{base}} = 0.3$ acts as a principled phase-transition boundary: it permits the maximum look-ahead budget for deterministic trajectories while ensuring that the drafted sequence remains a singularly dominant hypothesis. Truncation occurs immediately as the path enters high-entropy branching points where the joint reliability falls below this noise-floor. This theoretical lower bound profoundly aligns with our empirical grid search (see Fig. \ref{fig:threshold_latency}), where $\delta = 0.3$ yields the optimal equilibrium between drafting depth and acceptance ratio.

\section{Justification and Bound Analysis of the Expanded Draft Depth}
\label{sec:app_draft_depth}
In the main text, we propose expanding the maximum inference look-ahead budget to $H=2K$ within the Confidence-Gated Dynamic Drafting framework, governed by the calibrated confidence threshold $\delta$. To empirically validate this hyperparameter choice, we conduct an ablation study investigating the trade-off between generation efficiency (Latency) and the average accepted draft length ($\tau$) under varying maximum depth budgets. The results are summarized in Table \ref{tab:ablation_draft_depth}.\\
\begin{table}[h]
\centering
\small
\setlength{\tabcolsep}{8pt} 
\renewcommand{\arraystretch}{1.2} 
\caption{Comparison of efficiency and acceptance length across different look-ahead depths using the Confidence-Gated Dynamic Drafting strategy (basic look-ahead depth $K=9$).}
\label{tab:ablation_draft_depth}
\begin{tabular}{ccc}
\toprule
Look-ahead Depth & Latency$\downarrow$  & $\tau$$\uparrow$ \\
\hline
$K$ & 5.67 & 7.38  \\
$1.3K$ & 5.17 & 8.38  \\
$2K$ & \textbf{5.10} & 8.6  \\
$3K$ & 5.24 & \textbf{8.75} \\
\bottomrule
\end{tabular}
\end{table}

\noindent\textbf{Generalization Beyond Training Depth.} As demonstrated in Table \ref{tab:ablation_draft_depth}, restricting the inference draft depth to the training depth $K$ achieving an average acceptance length $\tau$ of only 7.38. By relaxing the upper bound to $1.3K$ and $2K$, the acceptance length monotonically increases to 8.38 and 8.60, respectively. This observation corroborates our hypothesis that the P-MTP curriculum equips the model with a generalized predictive capacity that extends well beyond its explicit training horizon $K$. Modulated by the threshold $\delta$, the model can confidently draft longer valid sequences during highly deterministic contexts.\\

\noindent\textbf{The Latency-Throughput Trade-off.} While a larger draft budget continuously increases the theoretical token yield, the end-to-end generation latency exhibits a distinct U-shaped trajectory. The optimal generation speed is achieved at $H=2K$ with a lowest latency of \textbf{5.10}. When the budget is aggressively expanded to $3K$, the latency degrades to 5.24, despite a marginal improvement in the acceptance length to 8.75.

This efficiency degradation occurs because the computational overhead of the drafting forward pass and the subsequent target verification eventually outweighs the diminishing returns of the extended acceptance length. Deep speculative trajectories are inherently more susceptible to cumulative uncertainty. Even if the local probabilities occasionally surpass $\delta$, the trailing tokens are highly prone to rejection by the primary verification mechanism. Consequently, evaluating these excessively long drafts incurs redundant FLOPs without proportionally contributing to the final output. Therefore, setting $H=2K$ strikes a favorable balance between performance and computational overhead, and our framework can achieve significant acceleration effects when deployed on multimodal foundational models of different scales.

\section{Defining the Training Look-ahead Horizon}
\label{sec:app_horizon}

In the main text, we demonstrated the scaling behavior of our method up to a look-ahead horizon of $K=9$. To comprehensively justify this architectural choice and understand the limits of our predictive modeling, we further investigate an extreme drafting horizon of $K=18$. The supplementary results are presented in Table \ref{tab:train_weight_horizon}.

Our empirical analysis reveals that expanding the look-ahead horizon beyond $K=9$ yields diminishing marginal returns and introduces significant optimization challenges. Specifically, we observe the following two key limitations when scaling to $K=18$:

\begin{table}[h]
\centering
\setlength{\tabcolsep}{3pt} 
\renewcommand{\arraystretch}{1.1} 
\caption{Additional results on training look-ahead horizon}
\label{tab:train_weight_horizon}
\begin{tabular}{llcccc}
\toprule
{{\makecell[l]{Weighting\\Scheme}}} & {{\makecell[l]{Drafting\\Setting}}} \quad\quad & {TEDS}$\uparrow$ & {Latency}$\downarrow$ & ${\tau}$$\uparrow$ & {Speedup}$\uparrow$ \\
\midrule
\rowcolor{shadecolor}
\multicolumn{6}{c}{{Base (NTP)}} \\
{-} & - &  85.09 & 26.71 & 1 & 1 \\
\midrule

\hline
\rowcolor{shadecolor}
\multicolumn{6}{c}{\textit{$K=9$}} \\
$\omega_k = 1$ & Fixed & 82.54 & 7.82 & 5.85 & 3.42 \\
$\omega_{k} = 0.2\times 0.8^k$  & Fixed & 85.33 & 5.95  & 7.05 & 4.49 \\
$\omega_{(t,k)}$ & Fixed & 84.78 &5.53 & 7.29 & 4.83 \\
$\omega_{(t,k)}$  & Dynamic & {84.78} &\textbf{5.10} & \textbf{8.60} & \textbf{5.24} \\

\hline
\rowcolor{shadecolor}
\multicolumn{6}{c}{\textit{$K=18$}} \\
$\omega_{k} = 1$  & Fixed & 77.18 & 17.32 & 1.53 & 1.54 \\
$\omega_{k} = 0.2\times 0.8^k$  & Fixed & 85.19 & 5.92 & 8.24 & 4.51 \\
$\omega_{(t,k)}$  & Fixed & 83.75 & 5.40 & 8.56& 4.95 \\
$\omega_{(t,k)}$  & Dynamic  & \textbf{83.75} & \textbf{5.20} & \textbf{9.00}& \textbf{5.13} \\
\bottomrule
\end{tabular}
\end{table}

\textbf{Efficiency Bottleneck:} Although increasing $K$ from 9 to 18 marginally improves the average acceptance length $\tau$ from 8.60 to 9.00 under our optimal dynamic weighting scheme $\omega_{(t,k)}$, this slight gain does not translate to faster wall-clock inference. In fact, the overall speedup drops from 5.24$\times$ to 5.13$\times$, and the latency increases. This performance regression indicates that the computational overhead introduced by generating and verifying 18 consecutive draft tokens outweighs the benefits of a slightly higher acceptance rate.

\textbf{Performance Degradation:} Pushing the predictive horizon too far exacerbates the difficulty of training the foundational model to accurately anticipate future contextual features. As shown in the results, even with our dynamic weighting strategy, the TEDS score for $K=18$ drops to 83.75, down from 84.78 at $K=9$. Furthermore, without appropriate distance-aware decay—specifically under the fixed $\omega_k = 1$ scheme—the generation quality collapses entirely to a TEDS of 77.18, alongside a minimal speedup of 1.54$\times$.

In conclusion, attempting to predict overly distant future tokens introduces excessive noise during training and imposes detrimental computational overhead during inference. Therefore, we establish $K=9$ as the optimal look-ahead horizon, which strikes the best balance between drafting capacity, generation fidelity, and end-to-end acceleration.

\end{document}